\pdfoutput=1

\documentclass[11pt]{article}

\usepackage[final]{acl}
\usepackage{tabularx}
\usepackage{times}
\usepackage{latexsym}
\usepackage{amsmath}
\usepackage[T1]{fontenc}

\usepackage[utf8]{inputenc}

\usepackage{microtype}

\usepackage{inconsolata}

\usepackage{graphicx}

\title{A Survey of QUD Models for Discourse Processing}

\author{Yingxue Fu \\
  School of Computer Science \\
  University of St Andrews\\
  Scotland, UK \\
  \texttt{fuyingxue321@gmail.com}}

\begin{document}
\maketitle
\begin{abstract}
Question Under Discussion (QUD), which is originally a linguistic analytic framework, gains increasing attention in the community of natural language processing over the years. Various models have been proposed for implementing QUD for discourse processing. This survey summarizes these models, with a focus on application to written texts, and examines studies that explore the relationship between QUD and mainstream discourse frameworks, including RST, PDTB and SDRT. Some questions that may require further study are suggested.  
\end{abstract}

\section{Introduction}
Mainstream discourse frameworks, such as Rhetorical Structure Theory (RST)~\citep{mann1988rhetorical}, Penn Discourse Treebank (PDTB)~\citep{webber-etal-2003-anaphora} and Segmented Discourse Representation Theory (SDRT)~\citep{asher2003logics}, are based on coherence relations. These relations are typically expressed with representative lexicons, such as \textit{Contrast} and \textit{Purpose}. 
Accordingly, discourse relation recognition is implemented as a multi-class classification task. 

In recent years, the Question Under Discussion (QUD) framework gains increasing attention as an alternative approach to discourse modelling, which is in line with the trend in converting NLP tasks into Question-Answering (QA) tasks~\citep{he-etal-2015-question, pyatkin-etal-2020-qadiscourse}. 

The QUD framework is originally a linguistic analytic framework for explaining pragmatic phenomena and information structural analysis~\citep{benz2017questions}. Early works by~\citet{von1989referential} and~\citet{van1995discourse} show how the framework can be applied for discourse modelling. The main idea is that discourse units, such as sentences, can be considered as answers to some implicit or explicit questions, called QUDs. One QUD may give rise to another QUD and some QUDs work together to answer a higher-level QUD. The organization of discourse units can be understood by the relationship between these QUDs.   

Although the conceptualization is simple, only in recent years, the implementation of the framework becomes a popular topic in discourse annotation and parsing. One reason is that reconstructing implicit QUDs for discourse units has been deemed an infeasible task because of the lack of constraints in the question generation process~\citep{riester2019constructing}. The second reason is that justification is needed for the reconstructed implicit QUDs, which makes the approach less favorable for analyzing written texts, where QUDs are implicit in most of the cases~\citep{benz2017questions}. Nevertheless, using free-form questions for discourse annotation is arguably simpler for lay annotators than using a fixed set of discourse relations predefined by experts, which are often abstract and ambiguous. Moreover, natural language generation (NLG) and QA tasks have been spurred by the development of large language models (LLMs), making the QUD approach to discourse modelling a potentially more cost-effective option than the other frameworks~\citep{ko-etal-2023-discourse}.  

A few challenges are discernible with implementing the QUD framework for discourse modelling. Similar to discourse frameworks based on coherence relations such as RST and PDTB, models that vary in the underlying theoretical assumptions have been proposed under the QUD framework, such as the QUD-tree approach by~\citet{riester2019constructing} and~\citet{de-kuthy-etal-2018-qud} and the expectation-driven approach by~\citet{westera-etal-2020-ted}. As these models are typically rooted in linguistic theories, it is difficult to understand the constraints and the properties of the representations obtained with the models. Meanwhile, 
the evaluation of QUD annotation and the automatic generation of QUDs are challenging due to the use of free-form questions, since a single QUD can be expressed in multiple ways.  

In this paper, we survey models for discourse processing proposed under the QUD framework to support researchers who are interested in applying this framework in downstream tasks 
or exploring discourse annotation and processing with the QUD framework~\citep{de-kuthy-etal-2018-qud, de-kuthy-etal-2020-towards}, or integrating different discourse frameworks~\citep{riester-etal-2021-combined, fu-2022-towards}. 

Compared to mainstream discourse frameworks, the body of literature on QUD is much smaller, particularly regarding the implementation. The studies covered in this survey were selected based on their influence in the field of computational linguistics, as reflected by follow-up works inspired by them. We began with seminal and high-impact studies and traced their references to develop a comprehensive understanding of the research landscape.

The contributions of this paper are summarized as follows:
\begin{enumerate}
   \item We identify three models for discourse processing under the QUD framework: QUD-tree approach, expectation-driven approach, and dependency-based approach. 
   \item We discuss the details and theoretical background of these models and compare them to highlight their respective properties. 
   \item We review studies on the relationship between mainstream discourse frameworks, including RST, PDTB and SDRT, and different QUD models.
\end{enumerate}


\section{QUD-Based Discourse Models} \label{existing-qud-studies}
Two general approaches to discourse processing can be identified within the QUD framework: the QUD-tree approach and the expectation-driven approach. The first approach is based on the theoretical proposal by~\citet{roberts2012information}, which uses trees to model the structure of QUDs. The second approach, named as such in the study by \citet{evaluating-expectation-driven}, is developed by \citet{onea2016potential}. In addition to the two canonical approaches,~\citet{ko-etal-2020-inquisitive} introduce another approach, which features inquisitive questions that can be anchored in the previous context (henceforth ``dependency-based QUD''). This approach is similar to the expectation-driven approach in some ways but involves the identification of the source that triggers the question in the previous context. 

\subsection{QUD-Tree Approach} \label{qud-tree-lr}
Early researchers, including~\citet{von1989referential} and~\citet{van1995discourse}, consider QUD as a principle for discourse structuring, where the relationships between sentences can be understood from the relationships between the QUDs they answer. Questions are raised one after another, and sentences either answer the questions or evoke further questions and answers in order to address the initial question.  
The final purpose  of the questioning process is to answer an overarching discourse question, called the \textit{Quaestio} of the text by~\citet{von1989referential}. 

This hierarchical relationship between questions may shape the choice of words and connectives. The restrictions imposed by QUDs on texts are studied by~\citet{von1989referential}. They refer to the incremental process of information unfolding in texts as \textit{referential movement}. To model this process, they categorize the different types of information within a proposition into five basic categories, called \textit{reference areas}, which include temporal properties, spatial properties, people involved, predicates (such as processes, states, and events), and modal properties. Within a proposition, information from the five areas is integrated into a whole, with information about predicates and people forming the inner core. The referential movement from one proposition to another is guided by the QUDs. For illustration, they use a simple narrative structure with a single characteristic QUD ``What happened to you?''. To answer this QUD, the propositions will describe events occurring during different time intervals t\textsubscript{i}. Therefore, the structure of the text can be delineated as ``Q1: What happened to you during the time interval t\textsubscript{1}'', ``Q2: What happened to you during the time interval t\textsubscript{2}'', and so on. 
Based on this structure of QUDs, they discuss whether references to entities in different reference areas should be maintained or adjusted during the referential movement.~\citet{von1989referential} recognize that texts with other structures exist, which cannot be characterized by one clear overarching QUD. These types of texts are weakly structured and coherence is established locally for a QUD. Furthermore,~\citet{von1989referential} highlight the existence of \textit{secondary structures}, where a QUD interrupts the flow of the main narrative structure. If secondary structures are developed extensively, they may give rise to a new high-level QUD. However,~\citet{von1989referential} perceive secondary structures negatively, because they believe that these structures do not contribute directly to the overall QUD of the text and involve violations of restrictions on referential movement. This negative view towards secondary structures is shared by~\citet{roberts2012information}.     

\citet{roberts2012information} posits that discourse is organized according to the strategies of inquiry used by discourse participants to share information and reach a common understanding about ``What is the way things are?''. This process involves a series of moves in which questions are posed and answered, leading to a reduction in ambiguities or indeterminacies. A question implies a set of alternatives, called the \textit{q-alternative} set of the question. A \textit{partial} answer to the question involves evaluating at least one item from the q-alternative set. A \textit{complete} answer, on the other hand, provides an evaluation of every item in the q-alternative set. Take the example from \citet{roberts2012information}:

\begin{center}
Q: Who did Mary invite?
\end{center}

If there are three people in the model of discourse, viz. P=\{Mary, Alice, Grace\}, the q-alternative set of the question \textit{Q} is: \{Mary invited Alice and Grace, Mary invited Alice but not Grace, Mary invited Grace but not Alice, Mary invited nobody\}\footnote{The case ``Mary invited Mary'' is ruled out, because ``invite'' is an irreflexive verb.}. Thus, the q-alternative set implies a partition of possible worlds. A complete answer suggests that one of the cells of the partition is chosen, such as ``Mary invited Alice but not Grace'', and the rest of the cells are discarded. In contrast, a partial answer would rule out some cells but require more moves to determine the final state. For example, an answer ``Mary invited Alice'' does not contain information about whether Grace is invited.

A \textit{focal alternative} set is defined for answers. When an answer ``Mary invited nobody'' is given to the question \textit{Q} above, if ``nobody'' is the focus, 
the focal alternative set would be: \{Mary invited Alice and Grace, Mary invited Alice but not Grace, Mary invited Grace but not Alice\}. In this case, the answer is \textit{congruent} with the question, because the focal alternative set of the answer matches the q-alternative set of the question. If the same answer is given but the focus is on ``Mary'', the focal alternative set would be: \{Alice invited nobody but Grace invited somebody, Alice and Grace invited nobody, Grace invited nobody but Alice invited somebody\}, which forms a counter-example to the idea that an answer should be congruent with the question. Therefore, in the theory by \citet{roberts2012information}, focus analysis, an aspect of information structural analysis, plays an important role.  

At the center of the theory by~\citet{roberts2012information} is the QUD stack, the bottom of which is the overarching question of the discourse, called the \textit{superquestion} by~\citet{roberts2012information}. \citet{roberts2012information} claims that when a question is accepted as a QUD and put on the QUD stack, the discourse participants will be committed to answering it until it is completely answered or until it is determined to be unanswerable. This is based on the assumption that rational discourse participants treat linguistic communication as a cooperative activity. Following Gricean maxims~\citep{grice1991studies}, the discourse participants will try to be relevant. 
Guided by the maxim of relevance, questions will be answered as soon as possible. The maxim of quantity requires the discourse participants to provide as much information as possible, favoring complete answers over partial ones. In the theory by~\citet{roberts2012information}, it is not required that questions on a QUD stack \textit{entail} those lower on the stack, where a question \textit{q\textsubscript{1}} entails a question \textit{q\textsubscript{2}} if and only if answering \textit{q\textsubscript{1}} yields a complete answer to \textit{q\textsubscript{2}}. Instead, a move should be \textit{relevant} to the immediate QUD, which means that the move should either introduce (at least) a partial answer to the QUD, when the move is an answer, or the move should be a part of the strategy to answer the QUD, if the move is a question\footnote{A question \textit{q\textsubscript{1}} \textit{contextually entails} a question \textit{q\textsubscript{2}} if and only if the union of the answer to \textit{q\textsubscript{1}} and the common ground when \textit{q\textsubscript{1}} is raised entails a complete answer to \textit{q\textsubscript{2}}. Being part of a strategy to answer a QUD denotes that a complete answer to the question will contextually entail a partial answer to the QUD.}. Answers that are irrelevant to the immediate QUDs are called ``non-sequiturs'', which reflect poor strategies and lack of commitment to the goal of communication, similar to the secondary structures discussed by~\citet{von1989referential}.

\citet{riester2019constructing} proposes a method for implementing the theory by \citet{roberts2012information}. 
The main idea is to introduce some constraints for reconstructing QUDs and make the process less dependent on phonological and syntactic analysis of texts, which is required in the approach demonstrated by~\citet{roberts2012information}. 

With the method by~\citet{riester2019constructing}, the first step involves discourse segmentation. Coordinated clauses, as well as verbal-phrase (VP) or determiner-phrase (DP) conjunctions or disjunctions, are segmented. When the resultant segments are incomplete, the elided materials are reconstructed so that a question can be inferred for each segment. After this step, the annotators are asked to infer QUDs for the segments and organize these QUDs into a hierarchical structure.   

The constraints for reconstructing QUDs include:
\begin{enumerate} 
\item Q-A-Congruence: QUDs must be answerable by the segments that they immediately dominate. 

This constraint is rooted in the notion of congruence proposed by~\citet{roberts2012information}, which requires that the q-alternative set matches the focal alternative set. However,~\citet{riester2019constructing}
does not follow this requirement strictly, only specifying that the interrogative part of a question is expected to be answered by the focus of the segment.

\item Q-Givenness: Implicit QUDs can only contain given materials. 

\citet{riester2019constructing} stresses that explicit QUDs are different from implicit ones because explicit QUDs can alter the information status of content by introducing new information into the discourse, whereas implicit QUDs are not meant to change the discourse, and therefore, they can only contain information from the previous context. 

\item Maximize-Q-Anaphoricity: Implicit QUDs should contain as much given material as possible. 

This constraint is introduced to ensure that reconstructed QUDs contain as many given details from the context as possible.

\item Back-to-the-Roots: This constraint concerns the attachment site of an incoming QUD and its answer.~\citet{riester2019constructing} adopts the right frontier constraint, which means that an incoming unit should be attached as low as necessary to allow for anaphoric reference but as high as possible to facilitate the conclusion of the current lower-level discourse and return to the main question of the discourse.
\end{enumerate}

At the same time, the method proposed by~\citet{riester2019constructing} deviates from the theoretical assumptions of~\citet{roberts2012information} in some aspects. First, the notion of ``being relevant'' is relaxed considerably. A discourse move is considered valid as long as a topical connection to the previous context can be established. Therefore, an incoming QUD and its answer are not bound to provide an answer to the QUDs on the stack. This choice is motivated by the suggestions by~\citet{hunter2015rhetorical} on the basis of comparing discourse structures formed by discourse relations and QUDs. However, as argued by~\citet{lee2024integrating}, it is difficult to formalize the notion of topical connection and this operation reduces the predictive power of the QUD model. The second deviation relates to the non-sequiturs. While \citet{roberts2012information} views information that does not contribute to answering the immediate QUD negatively, \citet{riester2019constructing} examines a broad range of such information under the notion of \textit{non-at-issue} content\footnote{The definition given by \citet{riester2019constructing}: An expression X of an utterance is \textit{non-at-issue} with respect to the current QUD of the utterance if and only if the deletion of X has no effect on the truth condition of the main proposition of the utterance.}, and believes that this kind of information represents new content. Even if it does not pertain to the immediate QUD, it may become the focus in subsequent discourse. Non-at-issue materials are annotated as a part of information structural analysis. The other categories covered in the annotation of information structure include focus (F), background (BG) and contrastive topic (CT)~\citep{de-kuthy-etal-2018-qud}. The focus refers to the constituent of the answer that addresses the current QUD, while the background denotes the remaining part that also addresses the QUD, but not in a salient status compared to the focus. The two parts form a focus domain. In the annotation model by~\citet{riester2019constructing}, contrastive topics refer to questions containing two interrogative parts, 
such as ``Who did what?''. These questions are typically answered by providing responses to subquestions about ``who'' and ``what'', respectively.   

\citet{de-kuthy-etal-2018-qud} report empirical results with the method by \citet{riester2019constructing}. 
In their experiments, two trained annotators were asked to annotate two sections of a transcript of an English interview. The Cohen's Kappa ($\kappa$) value is 0.52 on average, indicating moderate agreement. Results on annotating information structure show strong inter-annotator agreement.  

\begin{figure*}[t]
\begin{center}
\resizebox{\textwidth}{!}{
\includegraphics[width=1.0\textwidth]
  {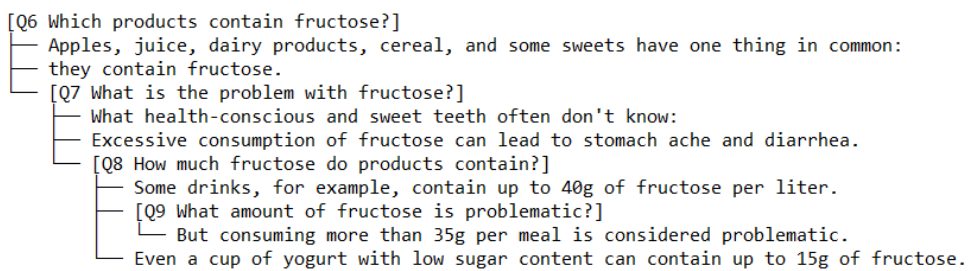} 
  }
  \caption{An illustration of annotation with the QUD-tree approach, from~\citet{shahmohammadi-etal-2023-encoding}. }
\label{qud-tree-example}
\end{center}
\vspace{-5mm}
\end{figure*}

An example of annotation using the method proposed by \citet{riester2019constructing} is illustrated in Figure~\ref{qud-tree-example}. 

The example below shows how information structure is annotated by \citet{de-kuthy-etal-2018-qud}. To improve consistency in the annotation, some heuristic rules are introduced. For instance, connectives are not annotated, which is illustrated by the case with ``and'', and pronouns are labeled BG, as shown by the annotation for ``one'':

\begin{quote} \label{info-str-example}
Q: {What kind of cars were there?}

A': [A]\textsubscript{BG} [red]\textsubscript{F} [one]\textsubscript{BG}

A'': and [a]\textsubscript{BG} [green]\textsubscript{F} [one]\textsubscript{BG}.
\end{quote}

\citet{de-kuthy-etal-2020-towards} investigate methods for the automatic generation of QUDs. Due to the lack of a large corpus annotated with QUDs and corresponding answer spans, they employ a rule-based method to create a corpus of triples consisting of a sentence, its associated question, and the phrase providing the answer within the sentence. This corpus is then used to train a neural question generation system. Experimental results demonstrate that a sequence-to-sequence model with an attention mechanism can generate questions comparable to those produced by the rule-based method.

\subsection{Expectation-Driven Approach} \label{expectation-lr}
As defined by~\citet{evaluating-expectation-driven}, an expectation-driven QUD model is one in which discourse participants anticipate the development of the discourse and use contextual cues to generate possible questions that subsequent sentences will answer. 

As one of the efforts investigating the implementation of this framework,~\citet{westera-etal-2020-ted} add a layer of QUD annotations with this approach to the English portion of TED-MDB (TED-Multilingual Discourse Bank)~\citep{zeyrek2020ted}. Their annotation effort centers on two questions: (a) what questions a discourse segment evokes, and (b) which question is answered in the subsequent discourse.  
To simplify the annotation task, annotators are presented with excerpts of a text, each excerpt comprising 18 sentences, and an annotator only works on 6 excerpts. The excerpts are shown incrementally to the annotators, and a probe point is set every two sentences, where the annotators are asked to enter a question evoked by the text up to that point, without knowledge about the following development of the discourse. Therefore, the questions should not have been answered up to that probe point. For questions that are evoked at previous two probe points but unanswered yet, the annotators are asked to indicate if the questions are answered at that probe point, based on a 5-point Likert scale, which encodes the degree to which the questions are answered, i.e., question answeredness. 

\begin{figure}[!h]
\begin{center}
\resizebox{\columnwidth}{!}{
\includegraphics[scale=0.45]
  {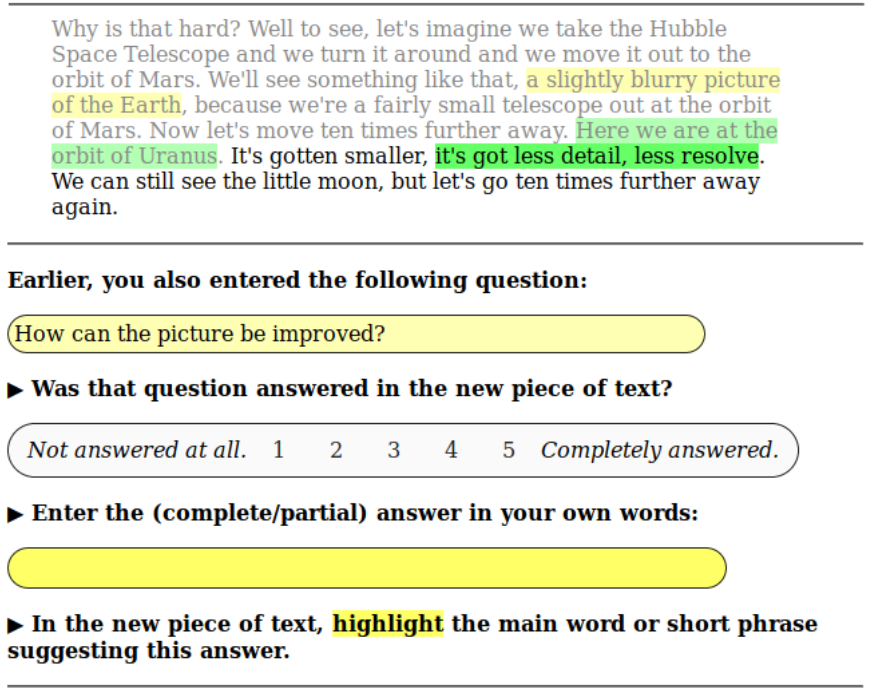} 
  }
  \caption{An illustration of checking the answeredness of a question, from~\citet{westera-etal-2020-ted}. }
\label{tedq-example}
\end{center}
\end{figure}

Figure~\ref{tedq-example} shows the annotation interface used by~\citet{westera-etal-2020-ted}. The annotators are also asked to highlight the words or spans in the sentence that form the most informative part of the answer. 

To ensure full coverage of the excerpts, probe points are alternated between participants. Since evaluating the relatedness of free-form questions automatically is non-trivial, an additional stage is introduced to manually estimate the semantic similarity of the evoked questions, serving as a measure of the reliability of questions evoked at each probe point. They detect a weak but statistically significant correlation between question reliability and question answeredness in the subsequent context.

\subsection{Dependency-Based Approach} \label{dependency-lr}
This approach was originally not targeted at discourse processing but focused on the QA task. \citet{ko-etal-2020-inquisitive} 
try to propose a new way of eliciting questions that reflect high-level understanding, in comparison with previous studies on factoid question answering, such as SQuAD~\citep{rajpurkar-etal-2018-know}. Annotators are shown one sentence at a time and they are asked to generate 0-3 questions, which are required to be grounded in a textual span of the current sentence. The questions are to be asked to increase the annotators' understanding of the text. Therefore, they call this type of questions \textit{inquisitive} questions. 
Owing to the challenging nature of this task, only the first five sentences of each news article are annotated. \citet{ko-etal-2020-inquisitive} propose three criteria for evaluating the generated questions, including (1) if the question is a complete and valid question; (2) if the question is related to the textual span; and (3) if the question has already been answered in the previous context. Although the purpose of the research differs from that of~\citet{westera-etal-2020-ted}, the two studies share commonalities in using free-form questions to represent expectation about the development of discourse at the local segment level. In addition, \citet{ko-etal-2020-inquisitive} explore automatic generation of inquisitive questions based on the dataset.

\citet{ko-etal-2022-discourse} extend the approach proposed by~\citet{ko-etal-2020-inquisitive} for discourse processing. Except for the first sentence, each sentence \textit{S} in a text is believed to be connected to a previous sentence by providing an answer to that sentence, which is called the \textit{anchor} (\textit{A}) of the question answered by \textit{S}. A free-form question (\textit{Q}) is used to describe the link between \textit{S} and \textit{A}. Therefore, discourse parsing can be formulated as a question of trying to identify \textit{A}, given {S} and the previous context \textit{C} of \textit{S}, and generating \textit{Q}. As free-form questions pose a challenge for evaluation, human judges are used to assess the similarity between the annotated questions. They find that 41.8\% of the questions provided by different annotators are considered highly similar, while an almost equal proportion, 40.7\%, are deemed semantically different. This approach combines dependency structure with QUD.  

\begin{figure}[t]
\begin{center}
\hbox{\hspace{-1.5em}{\includegraphics[scale=0.43]{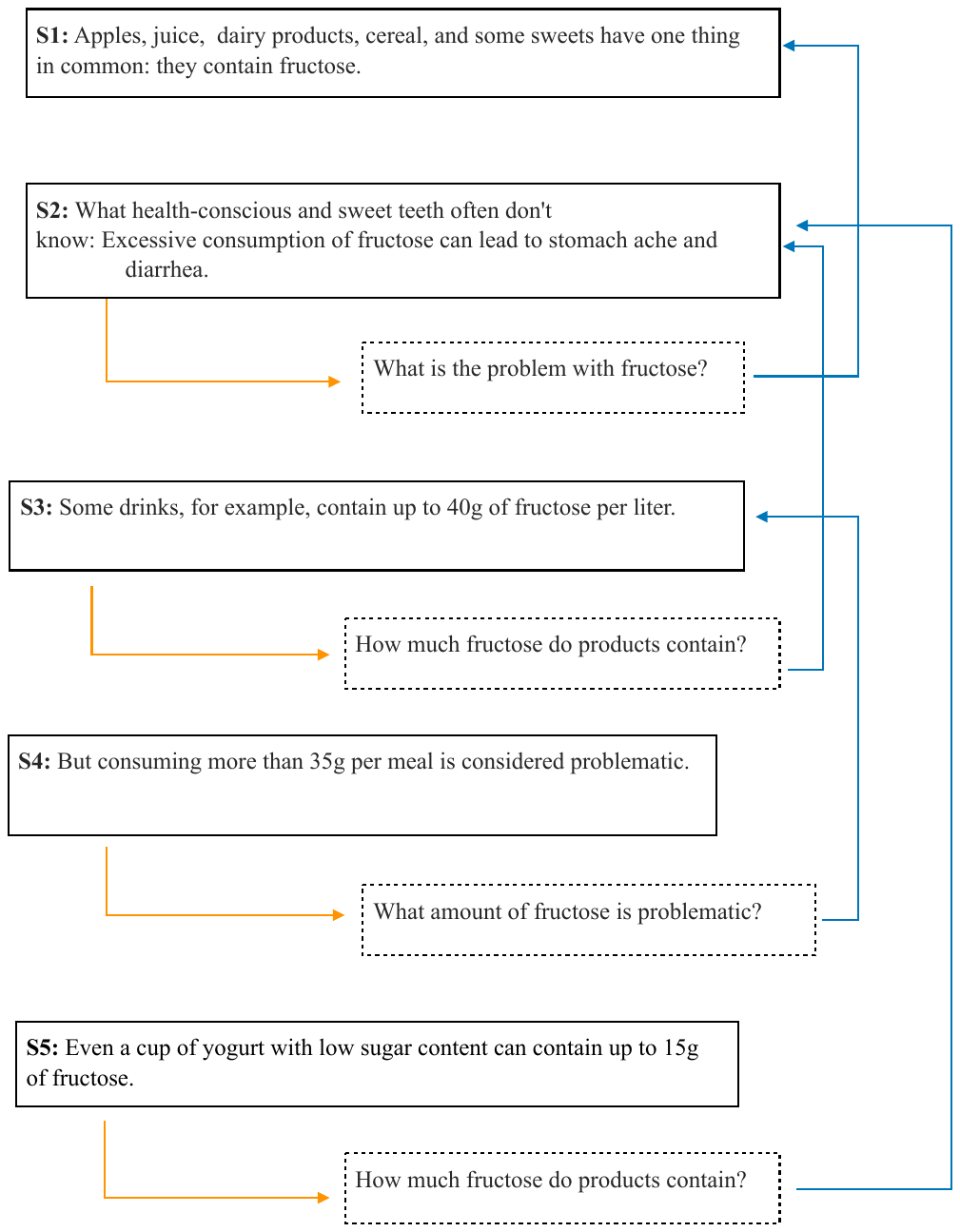} }}
\vspace{-3\baselineskip}
  \caption{An illustration of the analysis with the dependency-based QUD approach proposed by~\citet{ko-etal-2022-discourse}. Arrows in orange show the question-answer relationship, and arrows in blue show that the question is anchored in a previous sentence. For example, \textbf{S2} answers the question ``What is the problem with fructose?'', and the question is anchored in \textbf{S1}. As the questions are inferred, rather than being present in the original text, they are shown in dotted boxes.}
\label{qud-dep-example}
\end{center}
\vspace{-10mm}
\end{figure}

Figure~\ref{qud-dep-example} shows the discourse representation for the example in Figure~\ref{qud-tree-example} with the dependency-based QUD approach proposed by~\citet{ko-etal-2022-discourse}. It can be seen that the last sentence, \textbf{S5}, answers the same question as \textbf{S3} and the question is rooted in \textbf{S2}.

Based on the corpus created by~\citet{ko-etal-2022-discourse},~\citet{ko-etal-2023-discourse} develop a discourse dependency parser. To compare the annotations with RST,~\citet{ko-etal-2023-discourse} ask annotators to add RST relations to the corpus. They find that the questions annotated with the dependency-based QUD approach tend to be more fine-grained than RST relations, and more than one RST relation is possible where a question is annotated. This result is consistent with that of~\citet{hunter2015rhetorical}, although~\citet{hunter2015rhetorical} focus on the comparison between SDRT and QUD trees produced with the analysis of~\citet{roberts2012information}. As the corpus created by~\citet{ko-etal-2022-discourse} takes sentences as the basic unit, the comparison is made at the inter-sentential level. They also find that the QUD dependency structure differs from the dependency structure representation of RST trees, which is obtained with the method proposed by~\citet{hirao-etal-2013-single}. A similar observation is made by~\citet{shahmohammadi-etal-2023-encoding}. The reason is that RST distinguishes the salience of discourse segments linked by a relation, whereas QUD organizes the hierarchical structure based on the relationships between questions and subquestions.   

\citet{wu-etal-2023-qudeval} propose an evaluation suite for QUD dependency parsing. The criteria are rooted in the constraints of QUD reconstruction put forward by \citet{riester2018annotation} and \citet{de-kuthy-etal-2018-qud}. Given a sentence \textit{S} and the predicted anchor $\hat{\textit{A}}$ of the predicted question $\hat{\textit{Q}}$ answered by \textit{S}, the quality of $\hat{\textit{Q}}$ can be evaluated by four criteria: 1. if $\hat{\textit{Q}}$ is well-formulated linguistically and relevant to the content; 2. if the main part (\textit{focus} in terms of information structural analysis) of \textit{S} answers $\hat{\textit{Q}}$; 3. if $\hat{\textit{Q}}$ only contains concepts introduced or inferrable from the previous context; and 4. if $\hat{\textit{Q}}$ is closely related to $\hat{\textit{A}}$. Apart from the first criterion, which has only two values, `yes' and `no', each of the other criteria involves more nuanced evaluations. Moreover,~\citet{wu-etal-2023-qudeval} suggest that more than one $\hat{\textit{A}}$ is possible theoretically and operationally, and the results are heavily influenced by the specificity of questions. 

\citet{wu-etal-2024-questions} create a corpus containing annotations of the salience of evoked questions generated using the framework proposed by~\citet{ko-etal-2022-discourse}. A Likert scale from 1 to 5 is adopted to quantify the salience of questions. Multiple LLMs are employed to generate the questions, which are then evaluated by human annotators for their salience. Additionally, a subset of the questions is annotated with their answeredness in the following context, using a Likert scale from 0 to 3. They find that question salience is a statistically significant indicator of QUD, which is consistent with the observation of~\citet{westera-etal-2020-ted} who suggest that reliably evoked questions that are also answered in the subsequent context are likely to be the QUDs. However,~\citet{westera-etal-2020-ted} track whether a previously unanswered question is answered in the subsequent context at two consecutive probe points, while \citet{wu-etal-2024-questions} consider the entire subsequent context. This is because the model adopted by \citet{wu-etal-2024-questions} uses dependency links to represent the relations between sentences, which may involve long-distance connections. A question that remains to be studied with this approach is how to capture the hierarchical relationship between QUDs. As can be seen from Figure~\ref{qud-dep-example}, the relationship shown with dependency links is shallower than the QUD-tree approach shown in Figure~\ref{qud-tree-example}.

An overview of the approaches discussed in \ref{qud-tree-lr}, \ref{expectation-lr} and \ref{dependency-lr} is shown in Table~\ref{OverviewofQUDModels}. 

\begin{table*} \scriptsize
  \centering
  \begin{tabular}{p{0.07\textwidth}|p{0.07\textwidth}|p{0.07\textwidth}|p{0.2\textwidth}|p{0.15\textwidth}|p{0.08\textwidth}|p{0.07\textwidth}|p{0.06\textwidth}}
    \hline
     & Theoretical Foundation & Structural Assumption & Central Idea & Basic Unit & Non-at-issue Material & Number of QUDs per Edge & Edge-crossing \\
    \hline
   QUD-tree approach & ~\citet{roberts2012information} & a single tree & based on the constraints proposed by \citet{riester2019constructing}: Q-A-Congruence, Q-Givenness, Maximize-Q-Anaphoricity, Back-to-the-Roots & 
informational structural units, typically more fine-grained than sentences & captured, as long as a topical connection can be established &	1	& not allowed \\ \hline 

expectation-driven approach &~\citet{onea2016potential} &	not consider higher-level structure	& annotate questions evoked at a given discourse unit and expected to be answered in the following, not knowing the following discourse &	sentences or chunks of sentences&	not captured&	more than one &	possible, because questions evoked at a discourse unit will be tracked in the following few (2+) discourse units \\ \hline 

dependency-based approach & ~\citet{ko-etal-2022-discourse} & dependency structure	& treat each sentence as an answer to an implicit question triggered by a sentence in the previous context, which is called ``anchor'' & sentences &	not captured, unless a sentence is allowed to have no anchors	& theoretically and operationally can be more than 1	& possible \\ \hline
\end{tabular}
  \caption{An overview of existing QUD models.}
  \label{OverviewofQUDModels}
\end{table*}

\section{Benchmarks and Datasets}
Table~\ref{Overviewofbenchmarks} presents the benchmarks and datasets for different QUD models. 

\begin{table} \scriptsize
  \centering
  \begin{tabular}{p{1.0cm} p{2.0cm} p{3.40cm}}
    \hline
     & Systems Developed & Datasets \\ \hline 

QUD-tree approach & A full parser is not yet developed.~\citet{de-kuthy-etal-2020-towards} show a rule-based method of question generation. & 1. \citet{de-kuthy-etal-2018-qud} create a corpus comprising two sections of a transcript of an interview (English) (60 and 69 text segments, respectively) and a German radio interview (158 segments). 2.~\citet{shahmohammadi-etal-2023-encoding}
annotate 14 podcast transcripts and corresponding blog posts (German) (note that the segmentation in QUD annotation follows RST segmentation criteria). \\ \hline

Expectation-driven approach & \citet{wu-etal-2024-questions} employ instruction tuning of open-source LLMs. Agreement between results given by automatic methods and human annotations is 0.579 (Mean Absolute Error), 0.623 (Spearman), 0.417 (Macro F1) and 0.615 (Krippendorff's $\alpha$). & TED-Q~\citep{westera-etal-2020-ted} (the English portion of TED-MDB~\citep{zeyrek2020ted}.  \\ \hline 

Dependency-based approach & 1. \citet{ko-etal-2023-discourse} adopt a pipeline approach for parsing dependency structure of QUDs: the first step consists of predicting the anchor sentence and second step focuses on question generation and the generated questions are then reranked. 
2. \citet{suvarna-etal-2024-qudselect} use instruction-tuning to jointly predict the anchor sentences and the corresponding questions.  
& 1. DCQA~\citep{ko-etal-2022-discourse} contains 22,394 English human-generated question-answer pairs distributed across 606 English news articles. 2. QSALIENCE-data~\citep{wu-etal-2024-questions} is a corpus containing salience ratings and their natural language rationales for 1,766 inquisitive questions. 3. Regarding evaluation of QUDs: QUDEVAL~\citep{wu-etal-2023-qudeval} is a dataset comprising fine-grained evaluation of 2,190 QUDs over 51 news articles. 
\\ \hline

\end{tabular}
  \caption{An overview of existing benchmarks and datasets for each QUD model.}
  \label{Overviewofbenchmarks}
\end{table}

As pointed out by~\citet{riester2019constructing}, the reconstruction of QUDs typically does not rely on morphosyntactic signals, making the framework potentially cross-linguistically applicable. Works have shown that QUD is applicable to languages other than English, such as German~\citep{de-kuthy-etal-2018-qud, shahmohammadi-etal-2023-encoding} and Italian~\citep{de-kuthy-etal-2019-annotating}.

Similar to other mainstream discourse frameworks, such as RST and PDTB, QUD can be used for capturing higher-level linguistic information. As such, it is potentially useful for various NLP tasks that involve higher-level linguistic understanding. 
Applications already explored include facilitating narrative understanding~\citep{xu-etal-2024-fine}, where the dependency-based approach is adopted, conditional text generation~\citep{narayan-etal-2023-conditional}, which does not involve implementation of any of the existing QUD frameworks but only turns sentential information into question and answer pairs, similar to the first step of automatic question generation adopted by~\citet{de-kuthy-etal-2020-towards}, text simplification, specifically elaborative simplification~\citep{wu-etal-2023-elaborative}, and summarization~\citep{wu-etal-2024-questions}. 


\section{Relationship Between QUD and Discourse Frameworks}

This section reviews existing studies that explore the relationship between QUD and mainstream discourse frameworks including RST, PDTB, and SDRT

\subsection{QUD and RST} \label{qud-rst-rel}
\citet{shahmohammadi-etal-2023-encoding} annotate a corpus of 28 German texts, which are podcast transcripts and their corresponding blog posts. The corpus contains annotations of QUD trees in parallel with RST-style annotations. In order to simplify the comparison between the two frameworks, they apply the segmentation criteria of RST when performing QUD annotation, but the other steps of QUD annotation are performed following the guidelines developed by~\citet{riester2018annotation}. They convert RST-style annotations into a format similar to QUD-style annotations and compare the structures. A variant of the PARSEVAL metric for evaluating constituency parsing is adopted to measure the similarity between the tree structures of RST and QUD quantitatively. They find that the structural similarity is 74\% on average\footnote{The results are obtained considering the leaves, which may lead to higher scores than when leaves are excluded.}, and the similarity for monologues is higher than for dialogues, even though monologues are longer. Qualitative comparison shows that RST and QUD have similar patterns in grouping discourse segments, although there are cases where they differ owing to different focuses of the two frameworks. 
In terms of specific relations, they find that it is not straightforward to represent \textit{Background}, \textit{Restatement}, \textit{Concession} and \textit{Contrast} relations with QUD. 

\citet{riester-etal-2021-combined} propose a method of mapping RST and the CCR framework~\citep{sanders1992toward, sanders1993coherence} onto the QUD framework, represented by the QUD-tree approach~\citep{riester2019constructing}. A text of political speech annotated with the three frameworks in parallel is taken as an example. As segmentation in QUD is determined based on information structure, some smaller textual units that can function as answers to QUDs are considered as valid segments. This makes segmentation in QUD more fine-grained than RST and CCR, which typically take clauses or sentences as discourse segments. However, they argue that the more fine-grained segments in QUD can be captured by relations in RST, such as \textit{Elaboration}, \textit{Restatement} and \textit{List}. In addition, they discuss how subordinating and coordinating discourse relations can be expressed with QUD structures. The case with subordinating relations is straightforward: a subordinating relation can be converted into a subordinating QUD structure. For coordinating relations such as \textit{Conjunction},  \textit{Disjunction}, \textit{List} and \textit{Joint}, one QUD node dominates the coordinating segments, each segment providing a partial answer to the overarching question. Relations including \textit{Contrast} and \textit{Sequence} require the representation with a higher-level question, which takes questions that each segment answers respectively as its children, similar to the approach introduced by~\citet{von1989referential}. 

\subsection{QUD and PDTB}
In the research by \citet{westera-etal-2020-ted}, during the annotation process, annotators are only shown excerpts of texts and question-answeredness is tracked for only two consecutive chunks. Therefore, the discourse relations captured tend to be local, similar to PDTB-style annotations. The corpus they used also contains PDTB-style annotations, which allows an investigation into the relationships between the two frameworks. As free-form questions cannot be categorized easily, such as questions starting with ``how'' and ``what'', they only focus on \textit{why}-questions, which are potentially strongly correlated with causal relations. They find a statistically significant correlation between \textit{why}-questions and causal relations in PDTB including \textit{Cause}, \textit{Cause+Belief} and \textit{Purpose}. 

\subsection{QUD and SDRT}
\citet{hunter2015rhetorical} study the compatibility between QUD and SDRT. They highlight the fundamental differences between the two discourse models. If a stack is used for building a QUD tree, it is questions that are put on the stack during the process of tree construction. In contrast, during the process of constructing the hierarchical structure in the SDRT framework, it is discourse units, i.e., answers in the QUD framework, that are attached to the right frontier. The QUD tree approach proposed by~\citet{roberts2012information} follows a strict principle of organization based on questions and subquestions. A QUD is not popped off the stack until it is addressed. Therefore, the organization of the SDRT discourse graph does not necessarily mirror the QUD tree structure. Moreover, SDRT allows a node to have more than one parent, which is not possible under QUD. In addition, \citet{hunter2015rhetorical} show the challenges of representing some coordinating relations with QUD. Therefore, \citet{hunter2015rhetorical} reject a one-to-one correspondence between SDRT and QUD. Instead, they propose that CDUs, which group discourse units based on a common topic, have a similar planning function to QUDs, where questions are broken down into subquestions until the QUDs are manageable for discourse participants.

\section{Conclusion}
In this survey, three approaches for implementing the QUD framework for discourse representation are identified. Similar to the case with mainstream discourse frameworks represented by RST, PDTB and SDRT, these approaches exhibit varying focuses and capture different types of discourse information. 

While there are a few studies that explore the relationship between QUD-based discourse models and mainstream discourse frameworks, it remains an under-studied question whether QUDs represent the same type of discourse information as that encoded by discourse relations, for example, if \textit{why} questions encode causal relations consistently across the different frameworks and when \textit{why} questions are used for eliciting a \textit{Background} relation (if possible). Research on this question may provide insights on the strengths and weaknesses of different discourse frameworks. 
Moreover, 
it is still challenging to achieve high inter-annotator agreement, especially with the QUD-tree approach. In future work, a two-step approach similar to the method by~\citet{yung-etal-2019-crowdsourcing} can be adopted to control the QUD annotation process, where QUDs are elicited first and then categorized based on a predefined set of question templates that can be mapped unambiguously to discourse relations~\citep{pyatkin-etal-2020-qadiscourse}. Additionally, automatic QUD generation and application in downstream tasks also require further research.

\section{Limitations}
The theoretical background of the QUD-tree approach is discussed in details, because this approach is rooted in linguistic studies and the discussion may make it easier to understand later studies by~\citet{riester2019constructing}, which further forms the basis of the research by~\citet{wu-etal-2023-qudeval}. The expectation-driven approach is simpler and its theoretical foundation is not given much space.  

\section{Ethics Statement}
We do not foresee any ethical concerns with this survey. 

\bibliography{custom, anthology0, anthology1}


\end{document}